\documentclass[sigconf]{acmart}

\copyrightyear{2017}
\acmYear{2017}
\setcopyright{acmcopyright}
\acmConference{AISec'17}{November 3, 2017}{Dallas, TX, USA}
\acmPrice{15.00}
\acmDOI{10.1145/3128572.3140448}
\acmISBN{978-1-4503-5202-4/17/11}

\newcommand{\bx}{\mathbf{x}}
\newcommand{\by}{\mathbf{y}}
\newcommand{\be}{\mathbf{e}}
\newcommand{\bv}{\mathbf{v}}
\DeclareMathOperator*{\argmin}{arg\,min}
\usepackage{subcaption}
\usepackage{algorithm, algorithmic,url}
\usepackage{hyphenat}
\usepackage{hyperref}
\begin{document}
\title{ZOO: Zeroth Order Optimization Based Black-box Attacks to Deep Neural Networks without Training Substitute Models} 

\author{Pin-Yu Chen}
\authornote{Pin-Yu Chen and Huan Zhang contribute equally to this work.}
\affiliation{%
  \institution{AI Foundations Group \\ IBM T. J. Watson Research Center}
  \city{Yorktown Heights} 
  \state{NY} 
  \postcode{10598}
}
\email{pin-yu.chen@ibm.com}

\author{Huan Zhang}
\authornotemark[1]
\authornote{This work is done during internship at IBM T. J. Watson Research Center}
\affiliation{%
  \institution{University of California, Davis}
  \city{Davis} 
  \state{CA} 
  \postcode{95616}
}
\email{ecezhang@ucdavis.edu}

\author{Yash Sharma}
\affiliation{%
  \institution{IBM T. J. Watson Research Center}
  \city{Yorktown Heights} 
  \state{NY} 
  \postcode{10598}
}
\email{Yash.Sharma3@ibm.com}

\author{Jinfeng Yi}
\affiliation{%
  \institution{AI Foundations Group \\ IBM T. J. Watson Research Center}
  \city{Yorktown Heights} 
  \state{NY} 
  \postcode{10598}
}
\email{jinfengy@us.ibm.com}

\author{Cho-Jui Hsieh}
\affiliation{%
  \institution{University of California, Davis}
  \city{Davis} 
  \state{CA} 
  \postcode{95616}
}
\email{chohsieh@ucdavis.edu}

\begin{abstract}
Deep neural networks (DNNs) are one of the most prominent technologies of our time, as they achieve state-of-the-art performance in many machine learning tasks, including but not limited to image classification, text mining, and speech processing. However, recent research on DNNs has indicated ever-increasing concern on the robustness to adversarial examples, especially for security-critical tasks such as traffic sign identification for autonomous driving. Studies have unveiled the vulnerability of a well-trained DNN by demonstrating the ability of generating barely noticeable (to both human and machines) adversarial images that lead to misclassification. Furthermore, researchers have shown that these adversarial images are highly transferable by simply training and attacking a substitute model built upon the target model, known as a black-box attack to DNNs.

Similar to the setting of training substitute models, in this paper we propose an effective black-box attack that also only has access to the input (images) and the output (confidence scores)  of a targeted DNN. However, different from leveraging attack transferability from substitute models, we propose \textbf{z}eroth \textbf{o}rder \textbf{o}ptimization (\textit{ZOO}) based attacks to directly estimate the gradients of the targeted DNN for generating adversarial examples. We use zeroth order stochastic coordinate descent along with dimension reduction, hierarchical attack and importance sampling techniques to efficiently attack black-box models. By exploiting zeroth order optimization, improved attacks to the targeted DNN can be accomplished, sparing the need for training substitute models and avoiding the loss in attack transferability. Experimental results on MNIST, CIFAR10 and ImageNet show that the proposed ZOO attack is as effective as the state-of-the-art white-box attack (e.g., Carlini and Wagner's attack) and significantly outperforms existing black-box attacks via substitute models.
\end{abstract}




\maketitle


\section{Introduction}
\label{sec_intro}
The renaissance of artificial intelligence (AI) in the past few years roots in the advancement of deep neural networks (DNNs). Recently, DNNs have become an essential element and a core technique for existing and emerging AI research, as DNNs have achieved state-of-the-art performance and demonstrated fundamental breakthroughs 
in many machine learning tasks that were once believed to be challenging \cite{lecun2015deep}. Examples include computer vision, image classification, machine translation, and speech processing, to name a few.    

Despite the success of DNNs, recent studies have identified that DNNs can be vulnerable to adversarial examples - a slightly modified image can be easily generated and fool a well-trained image classifier based on DNNs with high confidence \cite{szegedy2013intriguing,goodfellow2014explaining}. Consequently, the inherent weakness of lacking robustness to adversarial examples for DNNs brings out security concerns, especially for mission-critical applications which require strong reliability, 
including traffic sign identification for autonomous driving and malware prevention \cite{Evtimov2017robust,hu2017generating,hu2017black}, among others.    

Preliminary studies on the robustness of DNNs focused on an ``open-box'' (white-box) setting - they assume model transparency that allows full control and access to a targeted DNN for sensitivity analysis. By granting the ability of performing back propagation, a technique that enables gradient computation of the output with respect to the input of the targeted DNN, many standard algorithms such as gradient descent can be used to attack the DNN. In image classification, back propagation specifies the effect of changing pixel values on 
the confidence scores for image label prediction. 
Unfortunately, most real world systems do not release their internal configurations (including network structure and weights), so open-box attacks cannot be used in practice. 

Throughout this paper, we consider a practical ``black-box'' attack setting where one can access the input and output of a DNN but not the internal configurations.  In particular, we focus on the use case where a targeted DNN is an image classifier trained by a convolutional neural network (CNN), which takes an image as an input and produces a confidence score for each class as an output. Due to application popularity and security implications, image classification based on CNNs is currently a major focus and a critical use case for studying the robustness of DNNs.

\begin{figure}[t]
		\centering
		\begin{subfigure}[b]{1\linewidth}
			\includegraphics[width=\textwidth]{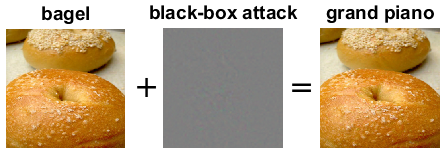}
			\caption{a ZOO black-box targeted attack example}
		\end{subfigure}%
        \vspace{2mm}
		\centering
		\begin{subfigure}[b]{1\linewidth}
			\includegraphics[width=\textwidth]{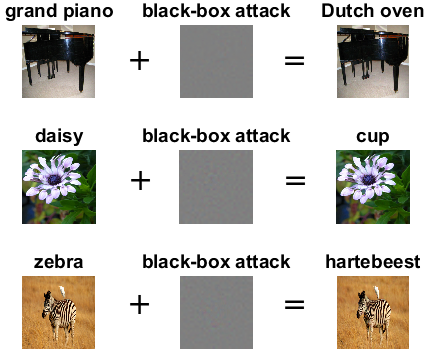}
			\caption{ZOO black-box untargeted attack examples}
		\end{subfigure}
		\caption{Visual illustration of our proposed black-box attack (ZOO) to sampled images from ImageNet. The columns from left to right are original images with correct labels, additive adversarial noises from our attack, and crafted adversarial images with misclassified labels.}
        \label{Fig_imagenet}
	\end{figure}

We consider two types of black-box attacks in this paper. Given a benign example with correct labeling, an \textit{untargeted} attack refers to crafting an  adversarial example leading to misclassification, whereas a \textit{targeted} attack refers to modifying the example in order to be classified as a desired class.  The effectiveness of our proposed black-box attack (which we call \textit{ZOO}) is illustrated in Figure \ref{Fig_imagenet}. The crafted adversarial examples from our attack not only successfully mislead the targeted DNN but also deceive human perception as the injected adversarial noise is barely noticeable. In an attacker's foothold, an adversarial example should be made as indistinguishable from the original example as possible in order to deceive a targeted DNN (and sometimes human perception as well). However, the best metric for evaluating the similarity between a benign example and a corresponding adversarial example is still an open question and may vary in different contexts.

In what follows, we summarize recent works on generating and defending adversarial examples for DNNs, and specify the ``black-box'' setting of training substitute models for adversarial attacks.

\subsection{Adversarial attacks and transferability}
\label{subsec_adv_attack}
\begin{figure}[t]
	\centering
	\includegraphics[width=.95\linewidth]{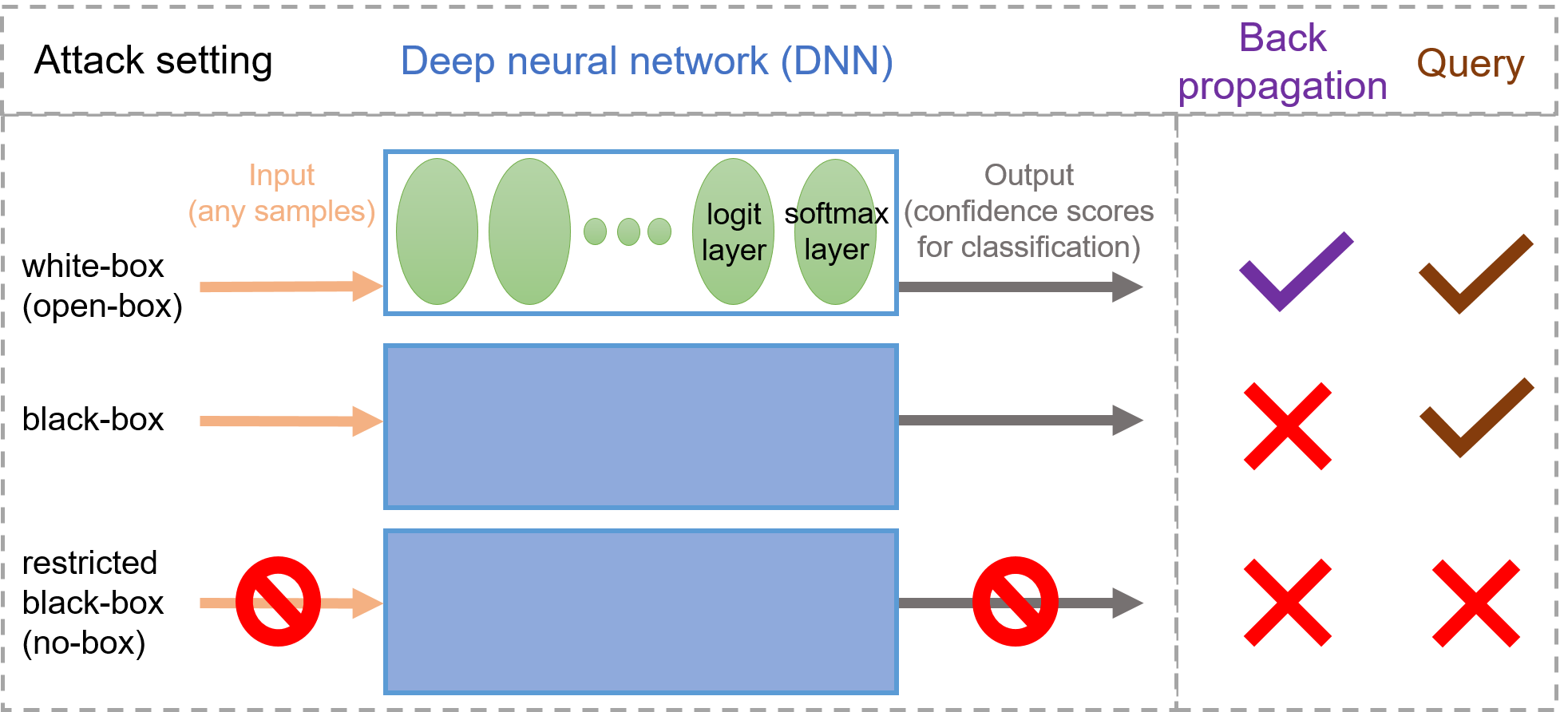}
	\caption{Taxonomy of adversarial attacks to deep neural networks (DNNs). ``Back propagation'' means an attacker can access the internal configurations in DNNs (e.g., performing gradient descent), and ``Query'' means an attacker can input any sample and observe the corresponding output. }	
	\label{Fig_attack_setting}
\end{figure}

We summarize four principal open-box methods developed for attacking image classification trained on DNNs as follows. 
\\
$\bullet$ \textbf{Fast gradient sign method (FGSM) \cite{goodfellow2014explaining}:} Originated from an $L_\infty$ constraint on the maximal distortion, FGSM uses the sign of the gradient from the back propagation on a targeted DNN to generate admissible adversarial examples. FGSM has become a popular baseline algorithm for improved adversarial example generation \cite{kurakin2016adversarial_ICLR,kurakin2016adversarial}, and it can be viewed as an attack framework based on first-order projected gradient descent \cite{madry2017towards}.
\\
$\bullet$   \textbf{Jacobian-based Saliency Map Attack (JSMA) \cite{papernot2016limitations}:} By constructing a Jacobian-based saliency map for characterizing the input-output relation of a targeted DNN, 
JSMA can be viewed as a greedy attack algorithm that iteratively modifies the most significant pixel based on the saliency map for crafting adversarial examples.  Each iteration,
JSMA recomputes the saliency map and uses the derivative of the DNN with respect to the input image as an indicator of modification for adversarial attacks.
In addition to image classification, JSMA has been applied to other machine learning tasks such as malware classification \cite{grosse2016adversarial}, and other DNN architectures such as recurrent neural networks (RNNs) \cite{papernot2016crafting}.
\\
$\bullet$ \textbf{DeepFool \cite{moosavi2016deepfool}:} Inspired from linear classification models and the fact that the corresponding separating hyperplanes indicate the decision boundaries of each class, DeepFool is an untargeted attack algorithm that aims to find the least distortion  (in the sense of Euclidean distance) leading to misclassification,
by projecting an image to the closest separating hyperplane. In particular, an approximate attack algorithm is proposed for DNNs in order to tackle the inherit nonlinearity for classification \cite{szegedy2013intriguing}.
\\
$\bullet$  \textbf{Carlini \& Wagner (C\&W) Attack \cite{carlini2017towards}:} The adversarial attack proposed by Carlini and Wagner is by far one of the strongest attacks.
They formulate targeted adversarial attacks as an optimization problem, take advantage of the internal configurations of a targeted DNN for attack guidance, and use the $L_2$ norm (i.e., Euclidean distance) to quantify the difference between the adversarial and the original examples. In particular, the representation in the logit layer (the layer prior to the final fully connected layer as illustrated in Figure \ref{Fig_attack_setting}) is used as an indicator of attack effectiveness. Consequently, the C\&W attack can be viewed as a gradient-descent based targeted adversarial attack driven by the representation of the logit layer of a targeted DNN and the $L_2$ distortion. The formulation of the C\&W attack will be discussed in detail in Section \ref{sec_black}. Furthermore, Carlini and Wagner also showed that their attack can successfully bypass 10 different detections methods designed for detecting adversarial examples \cite{carlini2017adversarial}.
\\
$\bullet$  \textbf{Transferability:} In the context of adversarial attacks, \textit{transferability} means that the adversarial examples generated from one model are also very likely to be misclassified by another model. In particular, the aforementioned adversarial attacks have demonstrated that their adversarial examples are highly transferable from one DNN at hand to the targeted DNN. One possible explanation of inherent attack transferability for DNNs lies in the findings that DNNs commonly  have overwhelming generalization power and local linearity for feature extraction \cite{szegedy2013intriguing}. 
Notably, the transferability of adversarial attacks brings about security concerns for machine learning applications based on DNNs, as malicious examples may be easily crafted even when the exact parameters of a targeted DNN are absent. More interestingly, the authors in \cite{moosavi2016universal} have shown that a carefully crafted universal perturbation to a set of natural images can lead to misclassification of all considered images with high probability, suggesting the possibility of attack transferability from one image to another. Further analysis and justification of a universal perturbation is given in \cite{moosavi2017analysis}.

\setlength{\belowcaptionskip}{-5pt}

\subsection{Black-box attacks and substitute models}
While the definition of an open-box (white-box) attack to DNNs is clear and precise - having complete knowledge  and allowing full access to a targeted DNN, the definition of a ``black-box'' attack to DNNs may vary in terms of the capability of an attacker. In an attacker's perspective, a black-box attack may refer to the most challenging case where only benign images and their class labels are given, but
the targeted DNN is completely unknown, and one is prohibited from querying any information from the targeted classifier for adversarial attacks. This restricted setting, which we call a ``no-box'' attack setting, excludes  the principal adversarial attacks introduced in Section \ref{subsec_adv_attack}, as they all require certain knowledge and back propagation from the targeted DNN. Consequently, under this no-box setting  the research focus is mainly on the attack transferability from one self-trained DNN to a targeted but completely access-prohibited DNN.

On the other hand, in many scenarios an attacker does have the privilege to query a targeted DNN in order to obtain useful information for crafting adversarial examples. For instance, a mobile app or a computer software featuring image classification (mostly likely trained by DNNs) allows an attacker to input any image at will and acquire classification results, such as the confidence scores or ranking for classification. An attacker can then leverage the acquired classification results to design more effective adversarial examples to fool the targeted classifier. In this setting, back propagation for gradient computation of the targeted DNN is still prohibited, as back propagation requires the knowledge of internal configurations of a DNN that are not available in the black-box setting. However, the adversarial query process can be iterated multiple times until an attacker finds a satisfactory adversarial example. For instance,  the authors in \cite{liu2016delving} have  demonstrated a successful black-box adversarial attack to Clarifai.com, which is a black-box image classification system.

Due to its feasibility, the case where an attacker can have free access to the input and output of a targeted DNN while still being prohibited from performing back propagation on the targeted DNN has been called a practical black-box attack setting for DNNs \cite{papernot2017practical,carlini2017towards,papernot2016transferability,hu2017generating,hu2017black,liu2016delving}. For the rest of this paper, we also refer a black-box adversarial attack to this setting. For illustration, the attack settings and their limitations are summarized in Figure \ref{Fig_attack_setting}.
 It is worth noting  that under this black-box setting, existing attacking approaches tend to make use of the power of free query to train a  \textit{substitute model}~\cite{papernot2017practical,papernot2016transferability,hu2017generating}, which is a representative substitute of the targeted DNN. The substitute model can then be attacked using any white-box attack techniques, and the generated adversarial images are used to attack the target DNN. The primary advantage of training a substitute model is its total transparency to an attacker, and hence essential attack procedures for DNNs, such as back propagation for gradient computation, can be implemented on the substitute model for crafting adversarial examples. Moreover, since the substitute model is representative of a targeted DNN in terms of its classification rules, adversarial attacks to a substitute model are expected to be similar to attacking the corresponding targeted DNN. In other words, adversarial examples crafted from a substitute model can be highly transferable to the targeted DNN given the ability of querying the targeted DNN at will.
 

\subsection{Defending adversarial attacks}
One common observation from the development of security-related research is that attack and defense often come hand-in-hand, and one's improvement depends on the other's progress. 
Similarly, in the context of robustness of DNNs, more effective adversarial attacks are often driven by improved defenses, and vice versa. 
There has been a vast amount of literature on enhancing the robustness of DNNs. Here we focus on the defense methods that have been shown to be effective in tackling (a subset of) the adversarial attacks introduced in Section \ref{subsec_adv_attack} while maintaining similar classification performance for the benign examples.
Based on the defense techniques,  we categorize the defense methods proposed for enhancing the robustness of DNNs to adversarial examples as follows.
\\
$\bullet$ \textbf{Detection-based defense:} Detection-based approaches aim to differentiate an adversarial example from a set of benign examples using statistical tests or out-of-sample analysis. 
Interested readers can refer to recent works in \cite{huang2016safety,metzen2017detecting,feinman2017detecting,grosse2017statistical,xu2017feature,xu2017feature_ext} and references therein for details. In particular, \textit{feature squeezing} is shown to be effective in detecting adversarial examples by projecting an image to a confined subspace (e.g., reducing color depth of a pixel) to alleviate the exploits from adversarial attacks  \cite{xu2017feature,xu2017feature_ext}.
The success of detection-based approaches heavily relies on the assumption that the distributions of adversarial and benign examples are fundamentally distinct. However, Carlini and Wagner recently demonstrated that their attack (C\&W attack) can bypass 10 different detection algorithms designed for detecting adversarial examples \cite{carlini2017adversarial}, which challenges the fundamental assumption of detection-based approaches as the results 
suggest that the distributions of their adversarial examples and the benign examples
are nearly indistinguishable. 
\\
$\bullet$ \textbf{Gradient and representation masking:} As the use of gradients via back propagation on DNNs has been shown to be crucial to crafting adversarial examples, one natural defense mechanism is to hide the gradient information while training a DNN, known as \textit{gradient masking}. A typical example of gradient masking is the defense distillation proposed in \cite{papernot2016distillation}. The authors proposed to retrain a DNN using distillation \cite{hinton2015distilling} based on the original confidence scores for classification (also known as soft labels) and introduced the concept of ``temperature'' in the softmax step for gradient masking. An extended version has been proposed to enhance its defense performance by incorporating model-agnostic uncertainty into retraining \cite{papernot2017extending}. Although the C\&W attack has shown to be able to break defensive distillation \cite{carlini2017towards}, it is still considered as a baseline model for defending adversarial attacks. Another defense technique, which we call \textit{representation masking}, is inspired by the finding that in the C\&W attack the logit layer representation in DNNs is useful for adversarial attacks. As a result, representation masking aims to replace the internal representations in DNNs (usually the last few layers) with robust representations to alleviate adversarial attacks. For example, the authors in \cite{bradshaw2017adversarial} proposed to integrate DNNs with Gaussian processes and RBF kernels for enhanced robustness.
\\
$\bullet$ \textbf{Adversarial training:} The rationale behind adversarial training is that DNNs are expected to be less sensitive to perturbations (either adversarial or random) to the examples if these adversarial examples are jointly used to stabilize training, known as the \textit{data augmentation} method. Different data augmentation methods have been proposed to improve the robustness of DNNs. Interested readers can refer to the recent works in \cite{jin2015robust,zheng2016improving,zantedeschi2017efficient,tramer2017ensemble,madry2017towards} and the references therein. Notably, the defense model proposed in \cite{madry2017towards} showed promising results against adversarial attacks, including the FGSM and the C\&W attack. The authors formulated defense in DNNs as a robust optimization problem, where the robustness is improved by iterative adversarial data augmentation and retraining. The results suggest that a DNN can be made robust at the price of increased network capacity (i.e., more model parameters), in order to stabilize training and alleviate the effect of adversarial examples.

\subsection{Black-box attack using zeroth order optimization: benefits and challenges}
Zeroth order methods are derivative-free optimization methods, where only the zeroth order oracle (the objective function value $f(\bx)$ at any $\bx$) is needed during optimization process. By evaluating the objective function values at two very close points $f(\bx + h \bv)$ and $f(\bx - h \bv)$ with a small $h$, a proper gradient along the direction vector $\bv$ can be estimated. Then, classical optimization algorithms like gradient descent or coordinate descent can be applied using the estimated gradients. The convergence of these zeroth order methods has been proved in optimization literature~\cite{nesterov2011random,ghadimi2013stochastic,lian2016comprehensive}, and under mild assumptions (smoothness and Lipschitzian gradient) they can converge to a stationary point with an extra error term which is related to gradient estimation and vanishes when $h \rightarrow 0$.

Our proposed black-box attack to DNNs in Section \ref{sec_black} is cast as an optimization problem. It exploits the techniques from zeroth order optimization and therefore spares the need of training a substitute model for deploying adversarial attacks. Although it is intuitive to use zeroth order methods to attack a black-box DNN model, applying it naively can be impractical for large models. 
For example, the Inception-v3 network \cite{szegedy2016rethinking} takes input images with a size of $299 \times 299 \times 3$, and thus has $p=268,203$ variables (pixels) to optimize. To evaluate the estimated gradient of each pixel, we need to evaluate the model twice. To just obtain the estimated gradients of all pixels, $2p=536,406$ evaluations are needed. For a model as large as Inception-v3, each evaluation can take tens of milliseconds on a single GPU, thus it is very expensive to even evaluate all gradients once. For targeted attacks, sometimes we need to run an iterative gradient descent with hundreds of iterations to generate an adversarial image, and it can be forbiddingly expensive to use zeroth order method in this case.

In the scenario of attacking black-box DNNs, especially when the image size is large (the variable to be optimized has a large number of coordinates), a single step of gradient descent can be very slow and inefficient, because it requires estimating the gradients of all coordinates to make a single update. Instead, we propose to use a coordinate descent method to iteratively optimize each coordinate (or a small batch of coordinates). By doing so, we can accelerate the attack process by efficiently updating coordinates after only a few gradient evaluations. 
This idea is similar to DNN training for large datasets, where we usually apply stochastic gradient descent using only a small subset of training examples for efficient updates, instead of computing the full gradient using all examples to make a single update. Using coordinate descent, we update coordinates by small batches, instead of updating all coordinates in a single update as in gradient descent. Moreover, this allows us to further improve the efficiency of our algorithm by using carefully designed sampling strategy to optimize important pixels first. We will discuss the detailed algorithm in Section~\ref{sec_black}.


\vspace{-1em}
\subsection{Contributions}
We refer to the proposed attack method as black-box attacks using \textbf{z}eroth \textbf{o}rder \textbf{o}ptimization, or \textbf{ZOO} for short. Below we summarize our main contributions:
\\
$\bullet$ We show that a coordinate descent based method using only the zeroth order oracle (without gradient information) can effectively attack black-box DNNs. Comparing to the substitute model based black-box attack \cite{papernot2017practical}, our method significantly increases the success rate for adversarial attacks, and attains comparable performance to the state-of-the-art white-box attack (C\&W attack).
\\
$\bullet$  
In order to speed up the computational time and reduce number of queries for our black-box attacks to large-scale DNNs, 
we propose several techniques including attack-space dimension reduction, hierarchical attacks and importance sampling.
\\
$\bullet$  In addition to datasets of small image size (MNIST and CIFAR10), we demonstrate the applicability of our black-box attack model 
to a large DNN - the Inception-v3 model trained on ImageNet. Our attack is capable of crafting a successful adversarial image within a reasonable time, whereas the substitute model based black-box attack in 
\cite{papernot2017practical} only shows success in small networks trained on MNIST and is hardly scalable to the case of ImageNet.


\vspace{-0.25em}
\section{Related Work}
\label{sec_related}
The study of adversarial attacks roots in the need for understanding the robustness of state-of-the-art machine learning models \cite{barreno2006can,barreno2010security}. For instance, Biggio et el. proposed an effective attack to sabotaging the performance (test accuracy) of support vector machines (SVMs) by intelligently injecting adversarial examples to the training dataset \cite{Biggio2012poison}. Gradient based evasion attacks to SVMs and multi-layer perceptrons are discussed in~\cite{biggio2013evasion}. Given the popularity and success of classification models trained by DNNs,  in recent years there has been a surge in interest toward understanding the robustness of DNNs. A comprehensive overview of adversarial attacks and defenses for DNNs is given in Section \ref{sec_intro}. 

Here we focus on related work on the black-box adversarial attack setting for DNNs. 
As illustrated in Figure \ref{Fig_attack_setting}, the black-box setting allows free query from a targeted DNN but prohibits any access to internal configurations (e.g., back propagation), which fits well to the scenario of publicly accessible machine learning services (e.g., mobile apps, image classification service providers, and computer vision packages). Under this black-box setting, the methodology of current attacks concentrates on training a substitute model and using it as a surrogate for adversarial attacks \cite{papernot2017practical,carlini2017towards,papernot2016transferability,hu2017generating,hu2017black,liu2016delving}. In other words, a black-box attack is made possible by deploying a white-box attack to the substitute model. Therefore, the effectiveness of such black-box adversarial attacks heavily depends on the attack transferability from the substitute model to the target model. Different from the existing approaches, we propose a black-box attack via zeroth order optimization techniques. More importantly, the proposed attack spares the need for training substitute models 
by enabling a ``pseudo back propagation'' on the target model. Consequently, our attack can be viewed ``as if it was'' a white-box attack to the target model, and its advantage over current black-box methods can be explained by the fact that it avoids any potential loss in transferability from a substitute model. The performance comparison between the existing methods and our proposed black-box attack will be discussed in Section \ref{sec_perm}.


In principle, our black-box attack technique based on zeroth order optimization is a general framework that can be applied to any white-box attacks requiring back propagation on the targeted DNN. We note that all the effective adversarial attacks discussed in Section \ref{subsec_adv_attack} have such a requirement, as back propagation on a targeted DNN provides invaluable information for an attacker. Analogously, one can view an attacker as an optimizer and an adversarial attack as an objective function to be optimized. Back propagation provides first-order evaluation (i.e., gradients) of the objective function for the optimizer for efficient optimization.
For the purpose of demonstration, in this paper we proposed a black-box attack based on the formulation of the C\&W attack, since the (white-box) C\&W attack has significantly outperformed the other attacks discussed in Section \ref{subsec_adv_attack} in terms of the quality (distortion) of the crafted adversarial examples and attack transferability \cite{carlini2017towards}. Experimental results in Section \ref{sec_perm} show that our black-box version is as effective as the original C\&W attack but at the cost of longer processing time for implementing pseudo back propagation. We also compare our black-box attack with the black-box attack via substitute models in \cite{papernot2017practical}, which trains a substitute model as an attack surrogate based on Jacobian saliency map \cite{papernot2016limitations}.
Experimental results in Section \ref{sec_perm} show that our attack significantly outperforms \cite{papernot2017practical}, which can be explained by the fact that our attack inherits the effectiveness of the state-of-the-art C\&W attack, and also by the fact that zeroth order optimization allows direct adversarial attacks to a targeted DNN and hence 
our black-box attack does not suffer from any loss in attack transferability from a substitute model.

\section{ZOO: A Black-box Attack without Training Substitute Models}
\label{sec_black}
\subsection{Notation for deep neural networks}
As illustrated in Figure \ref{Fig_attack_setting}, since we consider the black-box attack setting where free query from a targeted DNN is allowed while accessing to internal states (e.g., performing back propagation) is prohibited, it suffices to use the notation $F(\bx)$ to denote a targeted DNN. Specifically, the DNN $F(\bx)$ takes an image $\bx \in \mathbb{R}^p$ (a $p$-dimensional column vector) as an input and outputs a vector $F(\bx) \in [0,1]^{K}$ of confidence scores for each class, where $K$ is the number of classes. The $k$-th entry $[F(\bx)]_k \in [0,1]$ specifies the probability of classifying $\bx$ as class $k$, and $\sum_{k=1}^K [F(\bx)]_k=1$.

In principle, our proposed black-box attack via zeroth order optimization (ZOO) can be applied to non-DNN classifiers admitting the same input and output relationship. However, since DNNs achieved state-of-the-art classification accuracy in many image tasks, in this paper we focus on the capability of our black-box adversarial attack to DNNs.

\subsection{Formulation of C\&W attack}
Our black-box attack is inspired by the formulation of the C\&W attack \cite{carlini2017towards}, which is one of the strongest white-box adversarial attacks to DNNs at the time of our work. Given an image $\bx_0$, let $\bx$ denote the adversarial example of $\bx_0$ with a targeted class label $t$ toward misclassification. The C\&W attack finds   the adversarial example $\bx$ by solving the following optimization problem:
\begin{align}
\label{eqn_CW_attack}
&\textnormal{minimize}_{\bx}~\|\bx -\bx_0 \|_2^2 + c \cdot f(\bx,t) \\
&\textnormal{subject to}~\bx \in [0,1]^p, \nonumber
\end{align}
where $\| \mathbf{v} \|_2=\sqrt{\sum_{i=1}^p v_i^2}$ denotes the Euclidean norm ($L_2$ norm) of a vector $\mathbf{v}=[v_1,\ldots,v_p]^T$, and $c>0$ is a regularization parameter.

The first term $\|\bx -\bx_0 \|_2^2$ in (\ref{eqn_CW_attack}) is the regularization used to enforce the similarity between the adversarial example $\bx$ and the image $\bx_0$ in terms of the Euclidean distance, since $\bx - \bx_0$ is the adversarial image perturbation of $\bx$ relative to $\bx_0$. The second term $c \cdot f(\bx,t)$ in (\ref{eqn_CW_attack}) is the loss function that reflects the level of unsuccessful adversarial attacks, and $t$ is the target class.  Carlini and Wagner compared several candidates for $ f(\bx,t)$ and suggested the following form for effective targeted attacks \cite{carlini2017towards}:
\begin{align}
\label{eqn_CW_f}
f(\bx,t)=\max \{  \max_{i \neq t} [Z(\bx)]_i -  [Z(\bx)]_t    ,-\kappa  \},
\end{align}
where $Z(\bx) \in \mathbb{R}^K$ is the logit layer representation  (\textit{logits}) in the DNN for $\bx$ such that $[Z(\bx)]_k$ represents the predicted probability that $\bx$ belongs to class $k$, and $\kappa \geq 0$ is a tuning parameter for attack transferability. Carlini and Wagner set $\kappa=0$ for attacking a targeted DNN, and suggested large $\kappa$ when performing transfer attacks. The rationale behind the use of the loss function in (\ref{eqn_CW_f}) can be explained by the softmax classification rule based on the logit layer representation; the output (confidence score) of a DNN $F(\bx)$ is determined by the softmax function:
\begin{align}
\label{eqn_softmax}
[F(\bx)]_k=\frac{\exp([Z(\bx)]_k)}{\sum_{i=1}^K \exp([Z(\bx)]_i)},~\forall~k \in \{1,\ldots,K\}.
\end{align}
Therefore, based on the softmax decision rule in (\ref{eqn_softmax}), $\max_{i \neq t} [Z(\bx)]_i -  [Z(\bx)]_t \leq 0$ implies that the adversarial example $\bx$ attains the highest confidence score for class $t$ and hence the targeted attack is successful. On the other hand,  $\max_{i \neq t} [Z(\bx)]_i -  [Z(\bx)]_t > 0$ implies that the targeted attack using $\bx$ is unsuccessful. The role of $\kappa$ ensures a constant gap between $[Z(\bx)]_t$ and $\max_{i \neq t} [Z(\bx)]_i$, which explains why setting large $\kappa$ is effective in transfer attacks.

Finally, the box constraint $\bx \in [0,1]^p$ implies that the adversarial example has to be generated from the valid image space. In practice, every image can satisfy this box constraint by
dividing each pixel value by the maximum attainable pixel value (e.g., 255). Carlini and Wagner remove the box constraint by  replacing 
$\bx$ with $\frac{1+\tanh{\mathbf{w}}}{2}$, where $\mathbf{w} \in \mathbb{R}^p$. By using this change-of-variable, the optimization problem in (\ref{eqn_CW_attack}) becomes an unconstrained minimization problem with $\mathbf{w}$ as an optimizer, and typical optimization tools for DNNs (i.e., back propagation) can be applied for solving the optimal $\mathbf{w}$ and obtain the corresponding adversarial example $\bx$.

\subsection{Proposed black-box attack via zeroth order stochastic coordinate descent}
The attack formulation using (\ref{eqn_CW_attack}) and (\ref{eqn_CW_f}) presumes a white-box attack because (i): the logit layer representation in (\ref{eqn_CW_f})  is an internal state information of a DNN; and (ii) back propagation on the targeted DNN is required for solving (\ref{eqn_CW_attack}). We amend our attack to the black-box setting by proposing the following approaches: (i) modify the loss function $f(\bx,t)$ in  (\ref{eqn_CW_attack}) such that it only depends on the output $F$ of a DNN and the desired class label $t$; and (ii) compute an {\it approximate} gradient using a finite difference method instead of actual back propagation on the targeted DNN, and solve the optimization problem via zeroth order optimization. We elucidate these two approaches below.
\\
$\bullet$ \textbf{Loss function $f(\bx,t)$ based on $F$:} Inspired by (\ref{eqn_CW_f}), we propose a new hinge-like loss function based on the output $F$ of a DNN, which is defined as
\begin{align}
\label{eqn_CW_f_black_box}
f(\bx,t)=\max \{  \max_{i \neq t} \log [F(\bx)]_i -  \log [F(\bx)]_t    ,-\kappa  \},
\end{align}
where $\kappa \geq 0$ and $\log 0$ is defined as $-\infty$. We note that $\log(\cdot)$ is a monotonic function such that for any $x,y \geq 0$, $\log y \geq \log x$ if and only if $y \geq x$. This implies that $\max_{i \neq t} \log [F(\bx)]_i -  \log [F(\bx)]_t \leq 0$ means $\bx$ attains the highest confidence score for class $t$. We find that the log operator is essential to our black-box attack since very often a well-trained DNN yields a skewed probability distribution on its output $F(\bx)$ such that the confidence score of one class significantly dominates the confidence scores of the other classes. The use of the log operator lessens the dominance effect while preserving the order of confidence scores due to monotonicity. Similar to (\ref{eqn_CW_f}), $\kappa$ in (\ref{eqn_CW_f_black_box})  ensures a constant gap between $\max_{i \neq t} \log [F(\bx)]_i$ and  $\log [F(\bx)]_t$.

For untargeted attacks, an adversarial attack is successful when
$\bx$ is classified as any class other than the original class label $t_0$. A similar loss function can be used (we drop the variable $t$ for untargeted attacks): 
\begin{align}
\label{eqn_CW_f_black_box_untargeted}
f(\bx)=\max \{  \log [F(\bx)]_{t_0} - \max_{i \neq t_0} \log [F(\bx)]_i,-\kappa  \},
\end{align}
where $t_0$ is the original class label for $\bx$, and $\max_{i \neq t_0} \log [F(\bx)]_i$ represents the most probable predicted class other than $t_0$. 
\\
$\bullet$ \textbf{Zeroth order optimization on the loss function:}
We discuss our optimization techniques for any general function $f$ used for attacks (the regularization term in \eqref{eqn_CW_attack} can be absorbed as a part of $f$).
We use the symmetric difference quotient~\cite{lax2014calculus} to estimate the gradient $\frac{\partial f(\bx)}{\partial \bx_i}$ (defined as $\hat{g}_i$):

\begin{align}
\label{eq:grad_estimate}
\hat{g}_i \coloneqq \frac{\partial f(\bx)}{\partial \bx_i} \approx 
\frac{f(\bx + h \be_i) - f(\bx - h \be_i)}{2h},
\end{align}
where $h$ is a small constant (we set $h=0.0001$ in all our experiments) and $\be_i$ is a standard basis vector with only the $i$-th component as 1. The estimation error (not including the error introduced by limited numerical precision) is in the order of $O(h^2)$. Although numerical accuracy is a concern, accurately estimating the gradient is usually not necessary for successful adversarial attacks.  One example is FGSM, which only requires the sign (rather than the exact value) of the gradient to find adversarial examples. Therefore, even if our zeroth order estimations may not be very accurate, they suffice to achieve very high success rates, as we will show in our experiments.

For any $\bx \in \mathbb{R}^p$, we need to evaluate the objective function $2p$ times to estimate gradients of all $p$ coordinates. 
Interestingly, with just one more objective function evaluation, we can also obtain the coordinate-wise Hessian estimate (defined as $\hat{h}_i$):
\begin{align}
\label{eq:hess_estimate}
\hat{h}_i \coloneqq \frac{\partial^2 f(\bx)}{\partial \bx_{ii}^2} \approx 
\frac{f(\bx + h \be_i) - 2 f(\bx) + f(\bx - h \be_i)}{h^2}.
\end{align}
Remarkably, since $f(\bx)$ only needs to be evaluated once for all $p$ coordinates, we can obtain the Hessian estimates without additional function evaluations. 

It is worth noting that stochastic gradient descent and batch gradient descent are two most commonly used algorithms
for training DNNs, and the C\&W attack \cite{carlini2017towards} also used gradient descent to attack a DNN in the white-box setting. 
Unfortunately, in the black-box setting, the network structure is unknown and the gradient computation via back propagation is prohibited. To tackle this problem,
a naive solution is applying \eqref{eq:grad_estimate} to estimate gradient, which requires $2p$ objective function evaluations. However, this naive solution is too expensive in practice. Even for an input image size of $64\times 64 \times 3$, one full gradient descent step requires $24,576$ evaluations, and typically hundreds of iterations may be needed until convergence. 
To resolve this issue, we propose the following coordinate-wise update, which only requires $2$ function evaluations for each step. 	
\\

\begin{algorithm}
\caption{Stochastic Coordinate Descent}
  \label{alg:scd}
  \begin{algorithmic}[1]
  \WHILE{not converged}
  \STATE Randomly pick a coordinate $i \in \{1, \ldots, p\}$
  \STATE Compute an update $\delta^*$ by approximately minimizing
  \begin{equation*}
  \argmin_\delta f(\bx + \delta \be_i)
  \end{equation*}
  \STATE Update $\bx_i \leftarrow \bx_i + \delta^*$
  \ENDWHILE
 \end{algorithmic}
\end{algorithm}
$\bullet$ \textbf{Stochastic coordinate descent: }
Coordinate descent methods have been extensively studied in optimization literature~\cite{bertsekas1999nonlinear}. At each iteration, 
one variable (coordinate) is chosen randomly and is updated by approximately minimizing the objective function along that coordinate 
(see Algorithm~\ref{alg:scd} for details). 
The most challenging part in Algorithm~\ref{alg:scd} is to compute the best coordinate update in step 3. After estimating the gradient and Hessian for $\bx_i$, we can use any first or second order method to approximately find the best $\delta$. In first-order methods, we found that ADAM~\cite{kingma2014adam}'s update rule significantly outperforms vanilla gradient descent update and other variants in our experiments, so we propose to use a zeroth-order coordinate ADAM, as described in Algorithm~\ref{alg:scd-adam}. We also use Newton's method with both estimated gradient and Hessian to update the chosen coordinate, as proposed in Algorithm~\ref{alg:scd-newton}. Note that when Hessian is negative (indicating the objective function is concave along direction $\bx_i$), we simply update $\bx_i$ by its gradient. We will show the comparison of these two methods in Section \ref{sec_perm}.
Experimental results suggest coordinate-wise ADAM is faster than Newton's method. 

\begin{algorithm}
\caption{ZOO-ADAM: Zeroth Order Stochastic Coordinate Descent with Coordinate-wise ADAM}
  \label{alg:scd-adam}
  \begin{algorithmic}[1]
  \REQUIRE{Step size $\eta$, ADAM states $M \in \mathbb{R}^p, v \in \mathbb{R}^p, T \in \mathbb{Z}^p$, ADAM hyper-parameters $\beta_1 = 0.9$, $\beta_2 = 0.999$, $\epsilon = 10^{-8}$}
  \STATE $M \leftarrow \mathbf{0}, v \leftarrow \mathbf{0}, T \leftarrow \mathbf{0}$ 
  \WHILE{not converged}
  \STATE Randomly pick a coordinate $i \in \{1, \cdots, p\}$
  \STATE Estimate $\hat{g}_i$ using \eqref{eq:grad_estimate}
  \STATE $T_i \leftarrow T_i + 1$
  \STATE $M_i \leftarrow \beta_1 M_i + (1 - \beta_1) \hat{g}_i$, \quad $v_i \leftarrow \beta_2 v_i + (1 - \beta_2) \hat{g}_i^2$
  \STATE $\hat{M}_i = M_i / (1 - \beta_1^{T_i})$, \quad $\hat{v}_i = v_i / (1 - \beta_2^{T_i})$
  \STATE $\delta^* = - \eta \frac{\hat{M}_i}{\sqrt{\hat{v}_i} + \epsilon}$
  \STATE Update $\bx_i \leftarrow \bx_i + \delta^*$
  \ENDWHILE
 \end{algorithmic}
\end{algorithm}

\begin{algorithm}
\caption{ZOO-Newton: Zeroth Order Stochastic Coordinate Descent with Coordinate-wise Newton's Method}
  \label{alg:scd-newton}
  \begin{algorithmic}[1]
  \REQUIRE{Step size $\eta$}
  \WHILE{not converged}
  \STATE Randomly pick a coordinate $i \in \{1, \cdots, p\}$
  \STATE Estimate $\hat{g}_i$ and $\hat{h}_i$ using \eqref{eq:grad_estimate} and \eqref{eq:hess_estimate}
  \IF{$\hat{h}_i \leq 0$} 
    \STATE $\delta^* \leftarrow - \eta \hat{g}_i $ 
  \ELSE 
    \STATE $\delta^* \leftarrow - \eta \frac{\hat{g}_i}{\hat{h}_i} $
  \ENDIF
  \STATE Update $\bx_i \leftarrow \bx_i + \delta^*$
  \ENDWHILE
 \end{algorithmic}
\end{algorithm}

Note that for algorithmic illustration we only update one coordinate for each iteration. In practice, to achieve the best efficiency of GPU, we usually evaluate the objective in batches, and thus a batch of $\hat{g}_i$ and $\hat{h}_i$ can be estimated. In our implementation we estimate $B=128$ pixels' gradients and Hessians per iteration, and then update $B$ coordinates in a single iteration. 

\subsection{Attack-space dimension reduction}
\label{sec:dim_rec}

We first define $\Delta \bx = \bx - \bx_0$ and $\Delta \bx \in \mathbb{R}^p$ to be the adversarial noise added to the original image $\bx_0$. Our optimization procedure starts with $\Delta \bx = 0$. For networks with a large input size $p$, optimizing over $\mathbb{R}^p$ (we call it \textit{attack-space}) using zeroth order methods can be quite slow because we need to estimate a large number of gradients. 

Instead of directly optimizing $\Delta \bx \in \mathbb{R}^p$, we introduce a dimension reduction transformation $D(\by)$ where $\by \in \mathbb{R}^m$, $range(D) \in \mathbb{R}^p$, and $m < p$. The transformation can be linear or non-linear. Then, we use $D(\by)$ to replace $\Delta \bx = \bx - \bx_0$ in~\eqref{eqn_CW_attack}:
\begin{align}
\label{eqn_CW_attack_dim_reduce}
&\textnormal{minimize}_{\by}~\| D(\by) \|_2^2 + c \cdot f(\bx_0 + D(\by),t) \\
&\textnormal{subject to}~\bx_0 + D(\by) \in [0,1]^p. \nonumber
\end{align}
The use of $D(\by)$  effectively reduces the dimension of attack-space from $p$ to $m$. Note that we do not alter the dimension of an input image $\bx$ but only reduce the permissible dimension of the adversarial noise.
A convenient transformation is to define $D$ to be the upscaling operator that resizes $\by$ as a size-$p$ image, such as the bilinear interpolation method\footnote{See the details at \url{https://en.wikipedia.org/wiki/Bilinear_interpolation}}. For example, in the Inception-v3 network $\by$ can be a small adversarial noise image with dimension $m=32 \times 32 \times 3$, while the original image dimension is $p=299 \times 299 \times 3$. Other transformations like DCT (discrete cosine transformation) can also be used. We will show the effectiveness of this method in Section~\ref{sec_perm}.

\begin{figure*}[t]
\centering
\includegraphics[width=.75\linewidth]{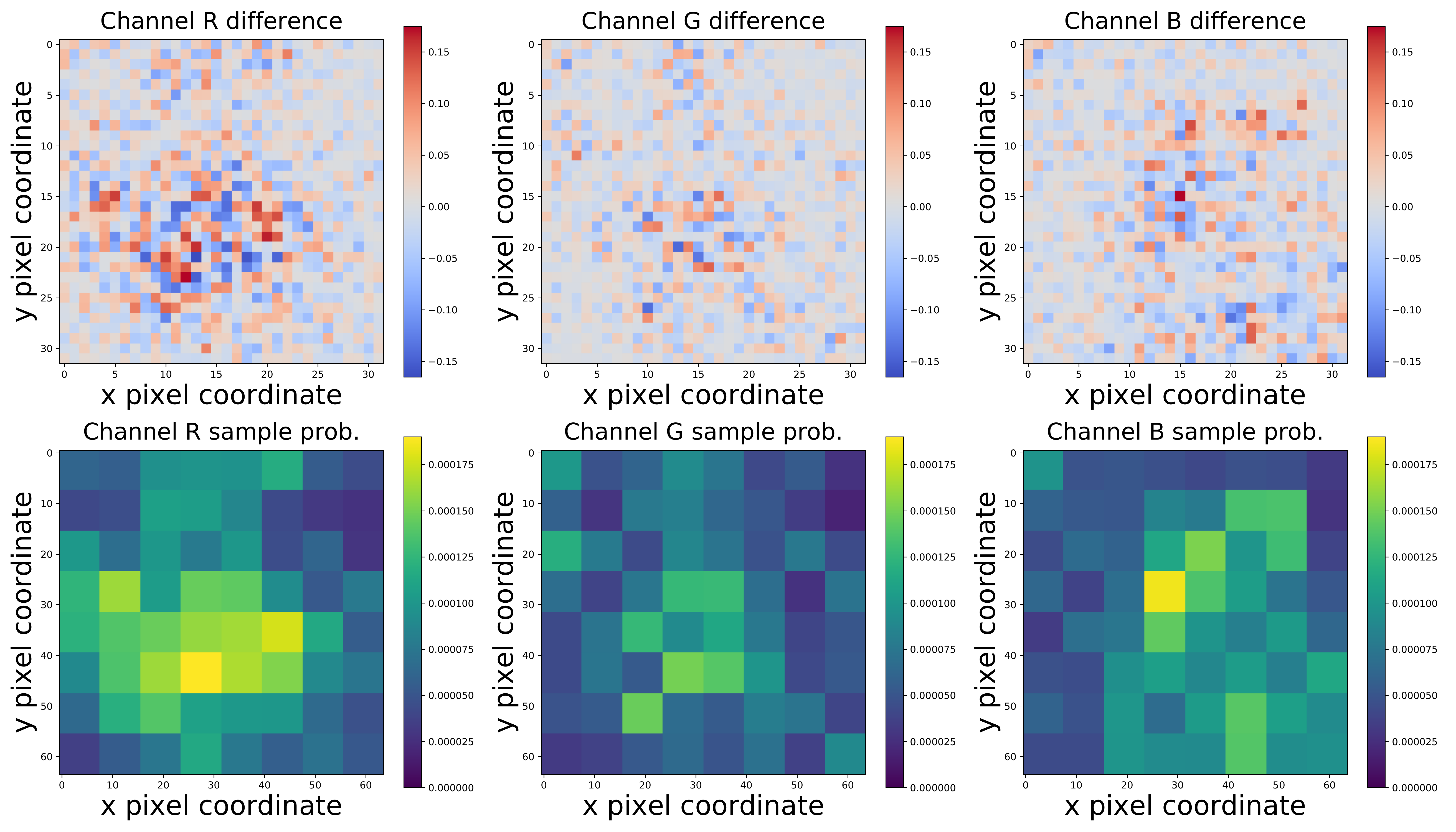}
		\caption{Attacking the bagel image in Figure~\ref{Fig_imagenet}~(a) with importance sampling.
        Top: Pixel values in certain parts of the bagel image have significant changes in RGB channels, and the changes in the R channel is more prominent than other channels. Here the attack-space is $32 \times 32 \times 3$. Although our targeted attack in this attack-space fails, its adversarial noise provides important clues to pixel importance. We use the noise from this attack-space to sample important pixels after we increase the dimension of attack-space to a larger dimension. Bottom: Importance sampling probability distribution for $64 \times 64 \times 3$ attack-space. The importance is computed by taking the absolute value of pixel value changes, running a $4 \times 4$ max-pooling for each channel, up-sampling to the dimension of $64 \times 64 \times 3$, and normalizing all values.
  }
        \label{Fig_importance}       
\end{figure*}

\subsection{Hierarchical attack}
\label{sec:hierarchy}

When applying attack-space dimension reduction with a small $m$, although the attack-space is efficient to optimize using zeroth order methods, a valid attack might not be found due to the limited search space. Conversely, if a large $m$ is used, a valid attack can be found in that space, but the optimization process may take a long time. Thus, for large images and difficult attacks, we propose to use a \textit{hierarchical attack} scheme, where we use a series of transformations $D_1, D_2 \cdots$ with dimensions $m_1, m_2, \cdots$ to gradually increase $m$ during the optimization process. In other words, at a specific iteration $j$ (according to the dimension increasing schedule) we set $\by_j = D_i^{-1}(D_{i-1}(\by_{j-1}))$ to increase the dimension of $\by$ from $m_{i-1}$ to $m_i$ ($D^{-1}$ denotes the inverse transformation of $D$).

For example, when using image scaling as the dimension reduction technique, $D_1$ upscales $\by$ from $m_1 = 32 \times 32 \times 3$ to $299 \times 299 \times 3$, and $D_2$ upscales $\by$ from $m_2 = 64 \times 64 \times 3$ to $299 \times 299 \times 3$. We start with $m_1 = 32 \times 32 \times 3$ variables to optimize with and use $D_1$ as the transformation, then after a certain number of iterations (when the decrease in the loss function is inapparent, indicating the need of a larger attack-space), we upscale $\by$ from $32 \times 32 \times 3$ to $64 \times 64 \times 3$, and use $D_2$ for the following iterations.

\subsection{Optimize the important pixels first}
\label{sec:importance}

One benefit of using coordinate descent is that we can choose which coordinates to update. Since estimating gradient and Hessian for each pixel is expensive in the black-box setting, we propose to selectively update pixels by using \textit{importance sampling}. For example, pixels in the corners or at the edges of an image are usually less important, whereas pixels near the main object can be crucial for a successful attack. Therefore, in the attack process we sample more pixels close to the main object indicated by the adversarial noise.

We propose to divide the image into $8 \times 8$ regions, and assign sampling probabilities according to how large the pixel values change in that region. We run a max pooling of the absolute pixel value changes in each region, up-sample to the desired dimension, and then normalize all values such that they sum up to 1. Every few iterations, we update these sampling probabilities according to the recent changes. In Figure~\ref{Fig_importance}, we show a practical example of pixel changes and how importance sampling probabilities are generated when attacking the bagel image in Figure \ref{Fig_imagenet} (a).

When the attack-space is small (for example, $32 \times 32 \times 3$), we do not use importance sampling to sufficiently search the attack-space. When we gradually increase the dimension of attack-space using hierarchical attack, incorporating importance sampling becomes crucial as the attack-space is increasingly larger. We will show the effectiveness of importance sampling in Section~\ref{sec_perm}.

\section{Performance Evaluation}
\label{sec_perm}
\subsection{Setup}

We compare our attack (ZOO) with Carlini \& Wagner's (C\&W) white-box attack~\cite{carlini2017towards} and the substitute model based black-box attack~\cite{papernot2017practical}. We would like to show that our black-box attack can achieve similar success rate and distortion as the white-box C\&W attack, and can significantly outperform the substitute model based black-box attack, while maintaining a reasonable attack time.

\begin{table*}[t]
\centering
\caption{MNIST and CIFAR10 attack comparison: ZOO attains comparable success rate and $L_2$ distortion as the white-box C\& W attack, and significantly outperforms the black-box substitute model attacks using FGSM  ($L_\infty$ attack) and the C\&W attack~\cite{papernot2017practical}. The numbers in parentheses in Avg. Time field is the total time for training the substitute model. For FGSM we do not compare its $L_2$ with other methods because it is an $L_\infty$ attack.
}
\label{fig:mnist-cifar}
\resizebox{1\textwidth}{!}{
\begin{tabular}{|l|c|c|c|c|c|c|}
\hline
                         & \multicolumn{6}{c|}{\textbf{MNIST}}                                                                               \\ \hline
                         & \multicolumn{3}{c|}{Untargeted}                 & \multicolumn{3}{c|}{Targeted}                          \\ \hline
                         &\ \ Success Rate\ \  &\ \  Avg. $L_2$ \ \   & \ \ Avg. Time (per attack)\ \  &\ \  Success Rate \ \       &\ \  Avg. $L_2$ \ \    &\ \  Avg. Time (per attack)\ \  \\ \hline
White-box (C\&W)         & 100 \%       & 1.48066 &        0.48 min                & 100 \%             & 2.00661  & 0.53 min               \\ \hline
Black-box (Substitute Model + FGSM) &       40.6 \%      & -       & 0.002 sec (+ 6.16 min)   & 7.48 \% & -  & 0.002 sec (+ 6.16 min)                     \\ \hline
Black-box (Substitute Model + C\&W) &       33.3 \%      & 3.6111  & 0.76 min (+ 6.16 min)    & 26.74 \%  & 5.272    & 0.80 min (+ 6.16 min)       \\ \hline
Proposed black-box (ZOO-ADAM) & 100 \%       & 1.49550 &       1.38 min                & 98.9 \%            & 1.987068 & 1.62 min               \\ \hline
Proposed black-box (ZOO-Newton) & 100 \%       & 1.51502 &     2.75 min              & 98.9 \%            & 2.057264 & 2.06 min               \\ \hline
                         & \multicolumn{6}{c|}{\textbf{CIFAR10}}                                                                            \\ \hline
                         & \multicolumn{3}{c|}{Untargeted}                 & \multicolumn{3}{c|}{Targeted}                          \\ \hline
                         & Success Rate & Avg. $L_2$   & Avg. Time (per attack) & Success Rate       & Avg. $L_2$    & Avg. Time (per attack) \\ \hline
White-box (C\&W)         & 100 \%       & 0.17980 &    0.20 min                    & 100 \%             & 0.37974  & 0.16 min               \\ \hline
Black-box (Substitute Model + FGSM)  &    76.1 \%         & -       & 0.005 sec (+ 7.81 min)   & 11.48 \% & -      & 0.005 sec (+ 7.81 min) \\ \hline
Black-box (Substitute Model + C\&W)  &    25.3 \%         & 2.9708  & 0.47 min (+ 7.81 min)    & 5.3 \% & 5.7439   & 0.49 min (+ 7.81 min) \\ \hline
Proposed Black-box (ZOO-ADAM) & 100 \%       & 0.19973 &   3.43 min                      & 96.8 \%            & 0.39879 & 3.95 min               \\ \hline
Proposed Black-box (ZOO-Newton) & 100 \%       & 0.23554 &     4.41 min                  & 97.0 \%            & 0.54226 & 4.40 min               \\ \hline
\end{tabular}
}
     \vspace*{1em}
\end{table*}

Our experimental setup is based on Carlini \& Wagner's framework\footnote{\url{https://github.com/carlini/nn_robust_attacks}} with our ADAM and Newton based zeroth order optimizer included. For substitute model based attack, we use the reference implementation (with necessary modifications) in CleverHans\footnote{\url{https://github.com/tensorflow/cleverhans/blob/master/tutorials/mnist_blackbox.py}} for comparison. For experiments on MNIST and CIFAR, we use a Intel Xeon E5-2690v4 CPU with a single NVIDIA K80 GPU; for experiments on ImageNet, we use a AMD Ryzen 1600 CPU with a single NVIDIA GTX 1080 Ti GPU. Our experimental code is publicly available\footnote{\url{https://github.com/huanzhang12/ZOO-Attack}}.
For implementing zeroth order optimization, we use a batch size of $B=128$; i.e., we evaluate 128 gradients and update 128 coordinates per iteration. In addition, we set $\kappa = 0$ unless specified.

\subsection{MNIST and CIFAR10}

{\bf DNN Model.} For MNIST and CIFAR10, we use the same DNN model as in the C\&W attack (\cite{carlini2017towards}, Table 1). For substitute model based attack, we use the same DNN model for \textit{both} the target model and the substitute model. If the architecture of a targeted DNN is unknown, black-box attacks based on substitute models will yield worse performance due to model mismatch.

\noindent
{\bf Target images.} For targeted attacks, we randomly select 100 images from MNIST and CIFAR10 test sets, and skip the original images misclassified by the target model. For each image, we apply targeted attacks to all 9 other classes, and thus there are 900 attacks in total. For untargeted attacks, we randomly select 200 images from MNIST and CIFAR10 test sets.

\noindent
{\bf Parameter setting.} For both ours and the C\&W attack, we run a binary search up to 9 times to find the best $c$ (starting from 0.01), and terminate the optimization process early if the loss does not decrease for 100 iterations. We use the same step size $\eta=0.01$ and ADAM parameters $\beta_1 = 0.9, \beta_2 = 0.999$ for all methods. For the C\&W attack, we run 1,000 iterations; for our attack, we run 3,000 iterations for MNIST and 1,000 iterations for CIFAR. Note that our algorithm updates far less variables because for each iteration we only update $128$ pixels, whereas in the C\&W attack all pixels are updated based on the full gradient in one iteration due to the white-box setting. Also, since the image size of MNIST and CIFAR10 is small, we do not reduce the dimension of attack-space or use hierarchical attack and importance sampling.
For training the substitute model, we use 150 hold-out images from the test set and run 5 Jacobian augmentation epochs, and set the augmentation parameter $\lambda = 0.1$. We implement FGSM  and the C\&W attack on the substitute model for both targeted and untargeted transfer attacks to the black-box DNN. For FGSM, the perturbation parameter $\epsilon = 0.4$, as it is shown to be effective in~\cite{papernot2017practical}. For the C\&W attack, we use the same settings as the white-box C\&W, except for setting $\kappa=20$ for attack transferability and using 2,000 iterations.

\noindent
{\bf Other tricks.} When attacking MNIST, we found that the change-of-variable via $\tanh$ can cause the estimated gradients to vanish due to limited numerical accuracy when pixel values are close to the boundary (0 or 1). Alternatively, we simply project the pixel values within the box constraints after each update for MNIST. But for CIFAR10, we find that using change-of-variable converges faster, as most pixels are not close to the boundary. 

\begin{figure*}[t]
		\centering
		\begin{subfigure}[b]{0.05\linewidth}
			\includegraphics[width=\textwidth]{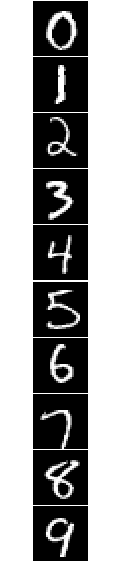}
			\caption{~}
		\end{subfigure}%
        		\hspace{5mm}
		\centering
		\begin{subfigure}[b]{0.235\linewidth}
			\includegraphics[width=\textwidth]{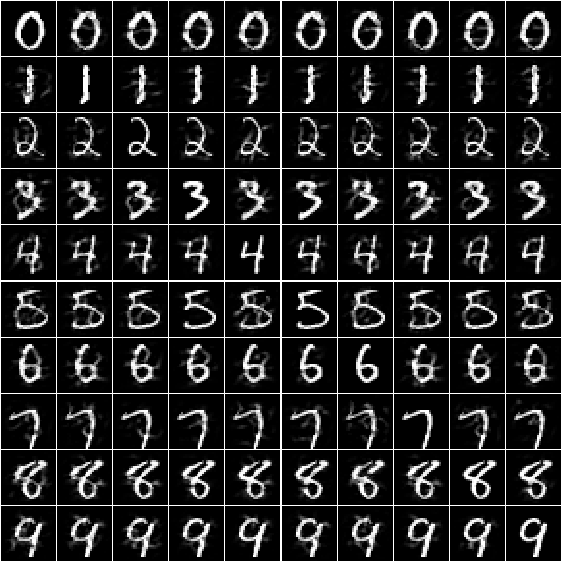}
			\caption{White-box C\&W attack}
		\end{subfigure}
                		\hspace{5mm}
		\centering                        
		\begin{subfigure}[b]{0.235\linewidth}
			\includegraphics[width=\textwidth]{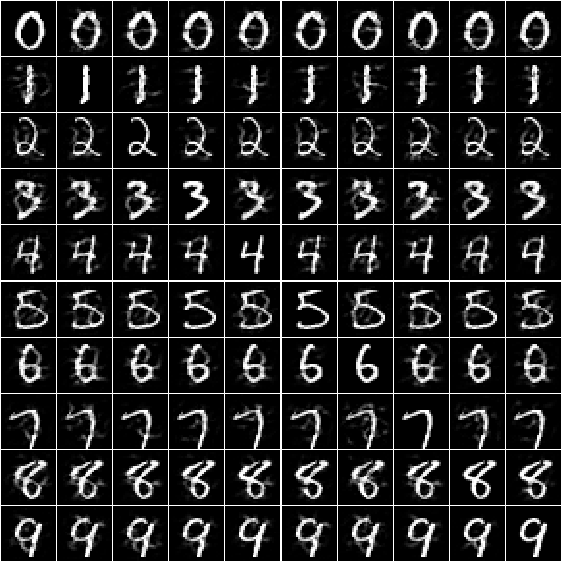}
			\caption{ZOO-ADAM black-box attack}
		\end{subfigure}        
                		\hspace{5mm}
		\centering                        
		\begin{subfigure}[b]{0.235\linewidth}
			\includegraphics[width=\textwidth]{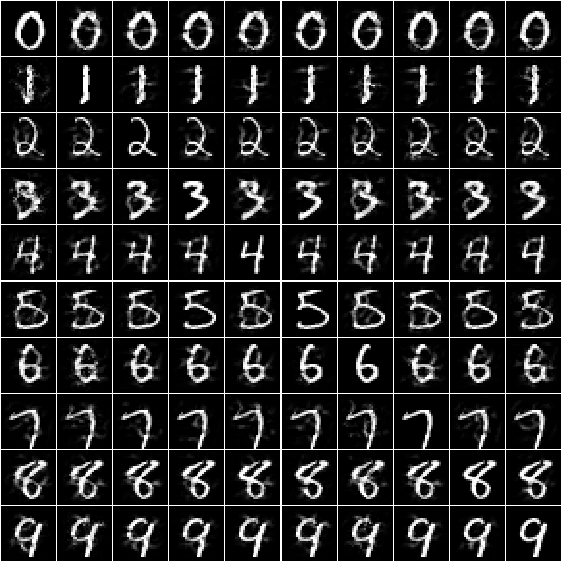}
		\caption{ZOO-Newton black-box attack}
		\end{subfigure}                
		\caption{Visual comparison of successful adversarial examples in MNIST. Each row displays crafted adversarial examples from the sampled images in (a). Each column in (b) to (d) indexes the targeted class for attack (digits 0 to 9). }
        \label{Fig_mnist}
        \vspace*{1em}
	\end{figure*}

\begin{figure*}[t]
		\centering
		\begin{subfigure}[b]{0.044\linewidth}
			\includegraphics[width=\textwidth]{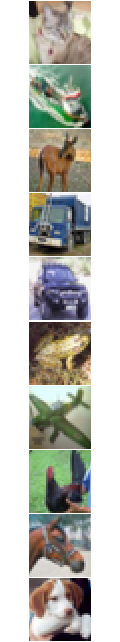}
			\caption{~}
		\end{subfigure}%
        		\hspace{5mm}
		\centering
		\begin{subfigure}[b]{0.235\linewidth}
			\includegraphics[width=\textwidth]{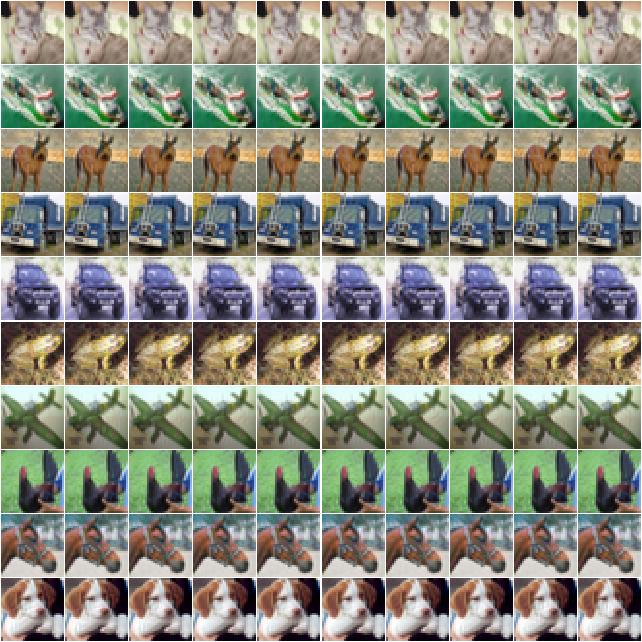}
			\caption{White-box C\&W attack}
		\end{subfigure}
                		\hspace{5mm}
		\centering                        
		\begin{subfigure}[b]{0.235\linewidth}
			\includegraphics[width=\textwidth]{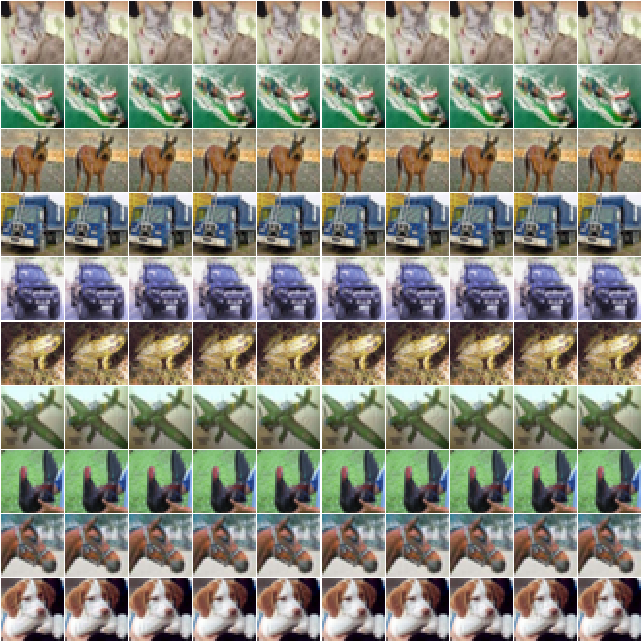}
			\caption{ZOO-ADAM black-box attack}
		\end{subfigure}        
                		\hspace{5mm}
		\centering                        
		\begin{subfigure}[b]{0.235\linewidth}
			\includegraphics[width=\textwidth]{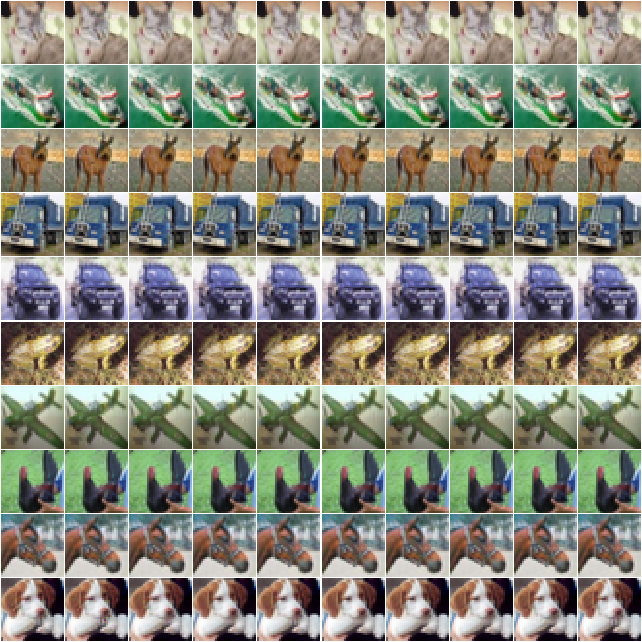}
		\caption{ZOO-Newton black-box attack}
		\end{subfigure}                
		\caption{Visual comparison of successful adversarial examples in CIFAR10. Each row displays crafted adversarial examples from the sampled images in (a). Each column in (b) to (d) indexes the targeted class for attack. }
        \label{Fig_cifar}
	\end{figure*}

\noindent
{\bf Results.}
As shown in Table~\ref{fig:mnist-cifar}, our proposed attack (ZOO) achieves nearly 100\% success rate. Furthermore, the $L_2$ distortions are also close to the C\&W attack, indicating our black-box adversarial images have similar quality as the white-box approach (Figures~\ref{Fig_mnist} and \ref{Fig_cifar}). Notably, our success rate is significantly higher than the substitute model based attacks, especially for targeted attacks, while maintaining reasonable average attack time. When transferring attacks from the substitute models to the target DNN, FGSM achieves better success rates in some experiments because it uses a relatively large $\epsilon=0.4$ and introduces much more noise than C\&W attack. We also find that ADAM usually works better than the Newton's method in terms of computation time and  $L_2$ distortion.

\subsection{Inception network with ImageNet}

Attacking a large black-box network like Inception-v3~\cite{szegedy2016rethinking} can be challenging due to large attack-space  and expensive model evaluation.
Black-box attacks via substitute models become impractical in this case, as a substitute model with a large enough capacity relative to Inception-V3 is needed, and a tremendous amount of costly Jacobian data augmentation is needed to train this model.
On the other hand, transfer attacks may suffer from lower success rate comparing to white-box attacks, especially for targeted attacks. Here we apply the techniques proposed in Section~\ref{sec:dim_rec}, \ref{sec:hierarchy} and \ref{sec:importance} to efficiently overcome the optimization difficulty toward effective and efficient black-box attacks.

\noindent
$\bullet$ \textit{\large \bf Untargeted black-box attacks to Inception-v3.}

\noindent
{\bf Target images.} We use 150 images from the ImageNet test set for untargeted attacks. To justify the effectiveness of using attack-space dimension reduction, we exclude small images in the test set and ensure that all the original images are at least $299 \times 299$ in size. We also skip all images that are originally misclassified by Inception-v3.

\setlength{\belowcaptionskip}{0pt}
\begin{table}[t]
\centering
\caption{Untargeted ImageNet attacks comparison. Substitute model based attack cannot easily scale to ImageNet. }
\label{tb:imgnet-untargeted}
\begin{tabular}{|l|c|c|}
\hline
                            & Success Rate & Avg. $L_2$ \\ \hline
White-box (C\&W)          & 100 \%       & 0.37310    \\ \hline
Proposed black-box (ZOO-ADAM) & 88.9 \%      & 1.19916    \\ \hline
 Black-box (Substitute Model) & N.A.     & N.A.    \\ \hline
\end{tabular}
\vspace{-0.5em}
\end{table}
\setlength{\belowcaptionskip}{-5pt}

\noindent
{\bf Attack techniques and parameters.}
We use an attack-space of only $32 \times 32 \times 3$ (the original input space is $299 \times 299 \times 3$) and do not use hierarchical attack. We also set a hard limit of $1,500$ iterations for each attack, which takes about 20 minutes per attack in our setup. In fact, during $1,500$ iterations, only $1500 \times 128 = 192,000$ gradients are evaluated, which is even less than the total number of pixels ($299 \times 299 \times 3 = 268,203$) of the input image. We fix $c=10$ in all Inception-v3 experiments, as it is too costly to do binary search in this case. For both C\&W and our attacks, we use step size 0.002.

	\begin{figure*}[htbp]
		\centering
		\begin{subfigure}[b]{.4\linewidth}
			\includegraphics[width=\textwidth]{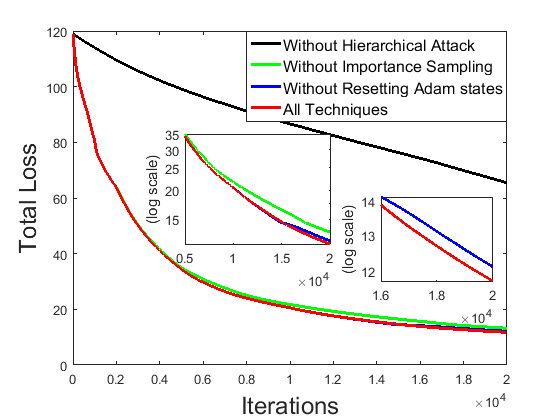}
		\end{subfigure}%
		\centering
		\begin{subfigure}[b]{.4\linewidth}
			\includegraphics[width=\textwidth]{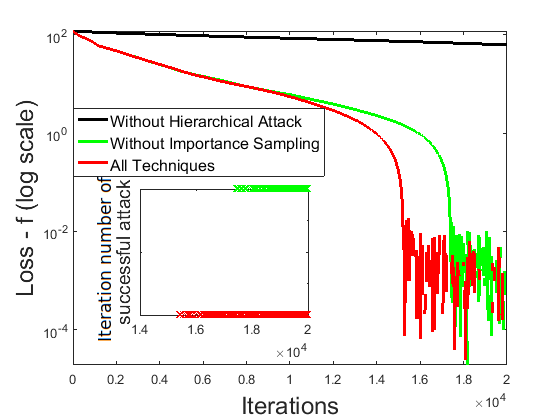}
		\end{subfigure}
		\caption{Left: total loss $\|\bx -\bx_0 \|_2^2 + c \cdot f(\bx,t)$ versus iterations. Right: $c \cdot f(\bx,t)$ versus iterations (log y-scale). When $c \cdot f(\bx,t)$ reaches 0, a valid attack is found. With all techniques applied, the first valid attack is found at iteration $15,227$. The optimizer then continues to minimize $\|\bx -\bx_0 \|_2^2$ to reduce distortion. In the right figure we do not show the curve without resetting ADAM states because we reset ADAM states only when $c \cdot f(\bx,t)$ reaches 0 for the first time.}
        \label{Fig_iter_loss}
	\end{figure*}

\noindent
{\bf Results.} We compare the success rate and average $L_2$ distortion between our ZOO attack and the C\&W white-box attack in Table~\ref{tb:imgnet-untargeted}. Despite running only 1,500 iterations (within 20 minutes per image) and using a small attack-space ($32 \times 32 \times 3$), our black-box attack achieves about 90\% success rate. The average $L_2$ distortion is about 3 times larger than the white-box attack, but our adversarial images are still visually indistinguishable (Figures~\ref{Fig_imagenet}).
The success rate and distortion can be further improved if we run more iterations.

\newpage
\noindent
$\bullet$ \textit{\large \bf Targeted black-box attacks to Inception-v3.}

For Inception-v3, a targeted attack is much more difficult as there are 1000 classes, and a successful attack means one can manipulate the predicted probability of any specified class. However, we report that using our advanced attack techniques, $20,000$ iterations (each with $B=128$ pixel updates) are sufficient for a \textit{hard} targeted attack.

\noindent
{\bf Target image.} We select an image (Figure~\ref{Fig_imagenet} (a)) for which our untargeted attack failed, i.e., we cannot even find an \textit{untargeted} attack in the $32 \times 32 \times 3$ attack-space, indicating that this image is hard to attack. Inception-v3 classifies it as a ``bagel'' with 97.0\% confidence, and  other top-5 predictions include ``guillotine'', ``pretzel'', ``Granny Smith'' and ``dough'' with 1.15\%, 0.07\%, 0.06\% and 0.01\% confidence. We deliberately make the attack even harder by choosing the target class as ``grand piano'', with original confidence of only 0.0006\%. 

\noindent
\textbf{Attack techniques.} 
We use attack-space dimension reduction as well as hierarchical attack. We start from an attack-space of $32 \times 32 \times 3$, and increase it to $64 \times 64 \times 3$ and $128 \times 128 \times 3$ at iteration 2,000 and 10,000, respectively. We run the zeroth order ADAM solver (Algorithm \ref{alg:scd-adam}) with a total of 20,000 iterations, taking about 260 minutes in our setup. Also, when the attack space is greater than $32 \times 32 \times 3$, we incorporate importance sampling, and keep updating the sampling probability after each iteration.

\noindent
\textbf{Reset ADAM states.} We report an additional trick to reduce the final distortion - reset the ADAM solver's states when a first valid attack is found during the optimization process. The reason is as follows.
The total loss consists of two parts: $l_1 \coloneqq c \cdot f(\bx,t)$ and $l_2 \coloneqq \|\bx -\bx_0 \|_2^2$. $l_1$ measures the difference between the original class probability $P_\text{orig}$ and targeted class probability $P_{\text{target}}$ as defined in (\ref{eqn_CW_f_black_box}). When $l_1 = 0$, $P_\text{orig} \leq P_{\text{target}}$, and a valid adversarial example is found. $l_2$ is the $L_2$ distortion. During the optimization process, we observe that before $l_1$ reaches 0, $l_2$ is likely to increase, i.e., adding more distortion and getting closer to the target class. After $l_1$ reaches 0 it cannot go below 0 because it is a hinge-like loss, and at this point the optimizer should try to reduce $l_2$ as much as possible while keeping $P_{\text{target}}$ only slightly larger than $P_\text{orig}$. However, when we run coordinate-wise ADAM, we found that even after $l_1$ reaches 0, the optimizer still tries to reduce $P_\text{orig}$ and to increase $P_{\text{target}}$, and $l_2$ will not be decreased efficiently. We believe the reason is that the historical gradient statistics stored in ADAM states are quite stale due to the large number of coordinates. Therefore, we simply reset the ADAM states after $l_1$ reaches 0 for the first time in order to make the solver focus on decreasing $l_2$ afterwards.

\setlength{\belowcaptionskip}{0pt}
\begin{table}[t]
\centering
\caption{Comparison of different attack techniques. ``First Valid'' indicates the iteration number where the first successful attack was found during the optimization process.}
\label{tb:imgnet-targeted}
\resizebox{.49\textwidth}{!}{
\begin{tabular}{|l|c|c|c|c|}
\hline
Black-box (ZOO-ADAM)              & Success? & First Valid & Final $L_2$ & Final Loss \\ \hline
All techniques         & Yes      & 15,227   & 3.425       & 11.735     \\ \hline
No Hierarchical Attack & No       & -                  & -           & 62.439     \\ \hline
No importance sampling & Yes      & 17,403   & 3.63486     & 13.216     \\ \hline
No ADAM state reset    & Yes      & 15,227   & 3.47935     & 12.111     \\ \hline
\end{tabular}
\vspace{-2em}
}
\end{table}
\setlength{\belowcaptionskip}{-5pt}

\noindent
\textbf{Results.}
Figure~\ref{Fig_iter_loss} shows how the loss decreases versus iterations, with all techniques discussed above applied in red; other curves show the optimization process without a certain technique but all others included. The black curve decreases very slowly, suggesting hierarchical attack is extremely important in accelerating our attack, otherwise the large attack-space makes zeroth order methods infeasible.
Importance sampling also makes a difference especially after iteration 10,000 -- when the attack-space is increased to $128 \times 128 \times 3$; it helps us to find the first valid attack over 2,000 iterations earlier, thus leaving more time for reducing the distortion.
The benefit of reseting ADAM states is clearly shown in Table~\ref{tb:imgnet-targeted}, where the final distortion and loss increase noticeably if we do not reset the states.

The proposed ZOO attack succeeds in decreasing the probability of the original class by over 160x (from 97\% to about 0.6\%) while increasing the probability of the target class by over 1000x (from 0.0006\% to over 0.6\%, which is top-1) to achieve a successful attack. Furthermore, as shown in Figures~\ref{Fig_imagenet},
the crafted adversarial noise is almost negligible and indistinguishable by human eyes. 

\section{Conclusion and Future Work}
\label{sec_con}
This paper proposed a new type of black-box attacks named ZOO to DNNs without training any substitute model as an attack surrogate. By exploiting zeroth order optimization for deploying pseudo back propagation on a targeted black-box DNN, experimental results show that our attack attains comparable performance to the state-of-the-art white-box attack (Carlini and Wagner's attack). In addition, our black-box attack significantly outperforms the substitute model based black-box attack in terms of attack success rate and distortion, as our method does not incur any performance loss in attack transferability. Furthermore, we proposed several acceleration techniques for applying our attack to large DNNs trained on ImageNet, whereas the substitute model based black-box attack is hardly scalable to a large DNN like Inception-v3. 

Based on the analysis and findings from the experimental results on MNIST, CIFAR10 and ImageNet, we discuss some potential research directions of our black-box adversarial attacks as follows.
\\
$\bullet$ Accelerated black-box attack: although our black-box attack spares the need for training substitution models and attains comparable performance to the white-box C\&W attack, in the optimization process it requires more iterations than a white-box attack due to additional computation for approximating the gradient of a DNN and for performing pseudo back propagation. In addition to the computation acceleration tricks proposed in Section \ref{sec_black}, a data-driven approach that take these tricks into consideration while crafting an adversarial example can make our black-box attack more efficient.
\\
$\bullet$ Adversarial training using our black-box attack: adversarial training in DNNs is usually implemented in a white-box setting - generating adversarial examples from a DNN to stabilize training and making the retrained model more robust to adversarial attacks. Our black-box attack can serve as an independent indicator of the robustness of a DNN for adversarial training. 
\\
$\bullet$ Black-box attacks in different domains: in this paper we explicitly demonstrated our black-box attack to image classifiers trained by DNNs. A natural extension is adapting our attack to other data types (e.g., speeches, time series, graphs) and different machine learning models and neural network architectures (e.g., recurrent neural networks). In addition, how to incorporate side information of a dataset  (e.g., expert knowledge) and existing adversarial examples (e.g., security leaks and exploits) into our black-box attack is worthy of further investigation.

\noindent
\section*{Acknowledgment}
The authors would like to warmly thank \nohyphens{Florian Tram\`{e}r} and\linebreak \nohyphens{Tsui-Wei Weng} for their valuable feedbacks and insightful comments. \nohyphens{Cho-Jui Hsieh} and Huan Zhang acknowledge the support of NSF via IIS-1719097. 

\bibliographystyle{ACM-Reference-Format}
\bibliography{ccs-sample}

\begin{figure*}[t]
\centering
\begin{subfigure}[b]{0.45\linewidth}
\includegraphics[width=\textwidth]{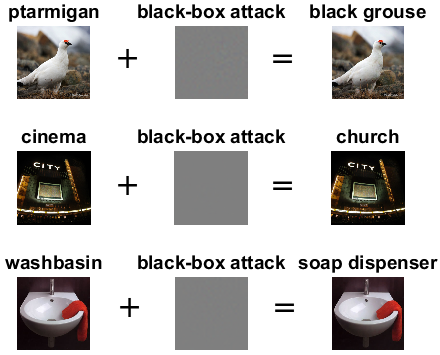}
\end{subfigure}%
\hspace{10mm}
\centering
\begin{subfigure}[b]{0.415\linewidth}
\includegraphics[width=\textwidth]{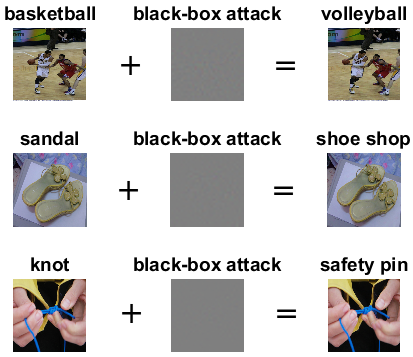}
\end{subfigure}
\caption{Additional visual illustration of black-box untargeted attacks using ZOO-ADAM, sampled from ImageNet test set. The columns from left to right are original images with correct labels, additive adversarial noises from our attack (gray color means no modification), and crafted adversarial images with misclassified labels. }
\label{Fig_imagenet_sup}
\end{figure*}

\section*{Appendix}
    
Figure \ref{Fig_imagenet_sup} displays some additional adversarial examples in ImageNet from our untargeted attack. \\


\end{document}